\documentclass[conference]{IEEEtran}
\makeatletter\if@twocolumn\PassOptionsToPackage{switch}{lineno}\else\fi\makeatother
\usepackage{bigdelim}

\usepackage{graphicx}
\usepackage[T1]{fontenc}
\usepackage[numbers,sort&compress]{natbib}
\usepackage{mathtools}
\usepackage{booktabs}
\usepackage{lineno,hyperref}
\usepackage{amsmath}
\usepackage{amsfonts}
\usepackage{amssymb}
\usepackage{amsmath}
\usepackage{multirow,morefloats,floatflt,cancel,tfrupee}
\usepackage{makecell}
\usepackage{tikz}
\usepackage{setspace}
\usepackage{color}
\usepackage{multirow}
\usepackage{adjustbox}
\usepackage{bbding}
\usepackage{physics}
\usepackage{colortbl}
\usepackage{xcolor}
\usepackage{pifont}
\usepackage[nointegrals]{wasysym}
\renewcommand\IEEEkeywordsname{Keywords}

\makeatletter

\let\old@ps@IEEEtitlepagestyle\ps@IEEEtitlepagestyle
\def\confheader#1{%
    \def\ps@IEEEtitlepagestyle{%
        \old@ps@IEEEtitlepagestyle%
        \def\@oddhead{\strut\hfill#1\hfill\strut}%
        \def\@evenhead{\strut\hfill#1\hfill\strut}%
    }%
    \ps@headings%
}
\makeatother

\confheader{%
    \parbox{17cm}{\centering 14th International Conference on Computer and Knowledge Engineering (ICCKE 2024), November 1-2, 2024, Ferdowsi University of Mashhad}
}

\begin{document}

        \title{Improving Dental Diagnostics: Enhanced Convolution with Spatial Attention Mechanism}

\author{\IEEEauthorblockN{Shahriar Rezaie \IEEEauthorrefmark{1},
Neda Saberitabar\IEEEauthorrefmark{2},
Elnaz Salehi\IEEEauthorrefmark{1}}\\

\IEEEauthorrefmark{1} Department of Electrical Engineering, Islamic Azad University, Qazvin, Iran \\
\IEEEauthorrefmark{2} Department of computer, software engineering, Islamic Azad University, Tehran, Iran\\
}


\maketitle

\begin{abstract}

Deep learning has emerged as a transformative tool in healthcare, offering significant advancements in dental diagnostics by analyzing complex imaging data. This paper presents an enhanced ResNet50 architecture, integrated with the SimAM attention module, to address the challenge of limited contrast in dental images and optimize deep learning performance while mitigating computational demands. The SimAM module, incorporated after the second ResNet block, refines feature extraction by capturing spatial dependencies and enhancing significant features. Our model demonstrates superior performance across various feature extraction techniques, achieving an F1 score of 0.676 and outperforming traditional architectures such as VGG, EfficientNet, DenseNet, and AlexNet. This study highlights the effectiveness of our approach in improving classification accuracy and robustness in dental image analysis, underscoring the potential of deep learning to enhance diagnostic accuracy and efficiency in dental care. The integration of advanced AI models like ours is poised to revolutionize dental diagnostics, contributing to better patient outcomes and the broader adoption of AI in dentistry.
\end{abstract}


\section{Introduction}

Deep learning, a subset of artificial intelligence, has emerged as a powerful tool in the field of healthcare, including dentistry \cite{lee2018detection}. By leveraging complex neural networks that can analyze vast amounts of data, deep learning algorithms have shown remarkable potential in enhancing the accuracy and efficiency of dental disease diagnosis. This technology can process and interpret various types of dental imaging, including X-rays, CT scans, and intraoral photographs, to detect and classify a wide range of dental conditions \cite{prajapati2017classification, schwendicke2019convolutional}.

The application of deep learning in dental diagnostics offers several advantages over traditional methods. First, it can identify subtle patterns and anomalies that might be overlooked by human practitioners, potentially leading to earlier detection of diseases such as dental caries, periodontal diseases, and oral cancers \cite{hung2020use}. Second, deep learning models can be trained on large datasets, allowing them to continuously improve their diagnostic accuracy over time \cite{tuzoff2019tooth}. This capability is particularly valuable in dentistry, where the interpretation of imaging can be subjective and varies among practitioners \cite{casalegno2019caries}.

Despite its promise, the integration of deep learning into dental practice faces several challenges. These include the need for large, high-quality datasets for training, concerns about the interpretability of AI-generated diagnoses, and the requirement for dental professionals to adapt to new technologies. However, as research progresses and more dental practices adopt these tools, deep learning is poised to become an invaluable asset in improving patient care, reducing diagnostic errors, and ultimately enhancing oral health outcomes.

To address the challenge of limited contrast in dental images and to leverage the advantages of deep learning while mitigating the high computational demands, we propose an enhanced ResNet50 architecture integrated with the SimAM attention module. Our approach involves incorporating the SimAM module after the second ResNet block, aiming to refine feature extraction and boost model performance in dental image analysis. The SimAM module is specifically designed to capture spatial dependencies within the images, enhancing significant features and suppressing less relevant ones. This integration aims to improve the network's ability to distinguish between characters and background, ultimately enhancing the classification accuracy and robustness of the model in dental applications.

 \begin{figure*}[!t]
 \centering
  \includegraphics[width=.8\textwidth]{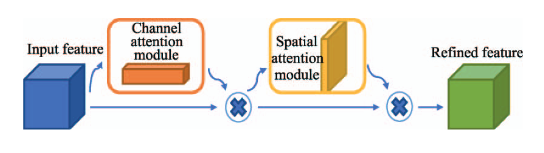}
  \caption{Convolution attention module.}\label{fig.ViT}
  \vspace{-4mm}
\end{figure*}

\section{Related Works}

\subsection{Convolutional Neural Networks}
Convolutional Neural Networks have been widely adopted in dental imaging analysis due to their ability to effectively process and analyze visual data. Lee et al. \cite{lee2018detection} demonstrated the efficacy of CNNs in detecting and diagnosing dental caries using panoramic radiographs, achieving high accuracy rates that rivaled those of experienced dentists. In periodontal disease assessment, Ekert et al. \cite{krois2019deep} applied CNNs to radiographic images for automated detection of periodontal bone loss, showing promising results in terms of sensitivity and specificity.
For tooth detection and numbering, a critical task in dental diagnostics, Tuzoff et al. \cite{tuzoff2019tooth} developed a CNN-based approach that achieved high accuracy on panoramic radiographs, potentially streamlining the diagnostic process. Chen et al. \cite{chen2019deep} extended this work by applying object detection techniques to periapical films, further demonstrating the versatility of CNNs in dental imaging tasks. In the realm of early caries detection, Casalegno et al. \cite{casalegno2019caries} utilized deep learning with near-infrared transillumination images, showcasing the potential of CNNs to enhance diagnostic capabilities beyond traditional radiographic methods.

\subsection{Transformers}
While CNNs have dominated the field of dental image analysis, Transformer architectures, originally developed for natural language processing tasks, have recently begun to make inroads in medical imaging, including dentistry. Manzari et al. \cite{manzari2024denunet} introduced a Transformer-based model for dental X-ray analysis, demonstrating superior performance in tooth segmentation and anomaly detection compared to traditional CNN approaches. The self-attention mechanism of Transformers allowed for better capture of global contextual information in dental radiographs.
Building on this work, Chen et al. \cite{chen2023cta} proposed a hybrid approach combining CNNs and Transformers for comprehensive dental disease diagnosis. Their model leveraged the local feature extraction capabilities of CNNs and the long-range dependency modeling of Transformers, resulting in improved accuracy across various dental pathologies. In a novel application, Zhang et al. \cite{duan20233d} utilized Vision Transformers for 3D dental mesh segmentation, showcasing the potential of these architectures in handling complex three-dimensional dental data.
Despite the promising results, research on Transformers in dental imaging is still in its early stages compared to CNNs. However, the ability of Transformers to handle sequence-to-sequence tasks and their potential for multi-modal learning (combining image and text data) suggest they may play an increasingly important role in comprehensive dental diagnostics and treatment planning systems in the future.

\section{METHODOLOGY}

Deep convolutional neural networks have achieved significant success in image classification, target detection, and image segmentation. With the advancement of deep learning, numerous excellent models have been developed. Commonly used image classification models in deep convolutional neural networks include LeNet, VGG, ResNet, and DenseNet. Given that the original image is a 3-channel image with minimal contrast between characters and the background in most cases, and deeper networks demand higher hardware resources for training, this paper selects ResNet50 as the base network and integrates SimAM into each residual block.

We propose an enhanced ResNet architecture incorporating the SimAM attention module after the second ResNet block to improve feature refinement and model performance in dental image analysis. The SimAM module is designed to capture spatial dependencies and enhance important features while suppressing less relevant ones.

\subsection{Revisiting ResNet}

ResNet, or Residual Network, has revolutionized the training of deep neural networks by introducing the concept of residual learning. This section revisits the formulation of ResNet, highlighting its key components and the motivation behind its design.

\subsection{Residual Learning}

ResNet's key principle is to learn residual functions with reference to the layer inputs rather than unreferenced functions. Let the required underlying mapping be $\mathcal{H}(x)$. ResNet allows layers to fit a residual mapping $\mathcal{F}(x) = \mathcal{H}(x) - x$, rather than simply mapping $x$ to $\mathcal{H}(x)$. The original function now becomes $\mathcal{H}(x) = \mathcal{F}(x) + x$.

\subsection{Residual Block}

A typical residual block can be expressed as:
\[
\mathbf{y} = \mathcal{F}(\mathbf{x}, \{W_i\}) + \mathbf{x}
\]
$\mathbf{x}$ and $\mathbf{y}$ are the input and output vectors of the layers under consideration. $\mathcal{F}(\mathbf{x}, \{W_i\})$ denotes the residual mapping to be learnt. For the simplest example, when the residual function has two layers, this may be expressed as:
\[
\mathcal{F}(\mathbf{x}, \{W_1, W_2\}) = W_2 \sigma(W_1 \mathbf{x})
\]
where $\sigma$ denotes the ReLU activation function.

\subsection{Bottleneck Design}

In deeper networks, a bottleneck design is used to reduce the number of parameters and computation cost. The bottleneck residual block is defined as:
\[
\mathbf{y} = W_3 \sigma(W_2 \sigma(W_1 \mathbf{x})) + \mathbf{x}
\]
where $W_1, W_2,$ and $W_3$ are 1x1, 3x3, and 1x1 convolutional layers, respectively.

\subsection{ResNet-50 Architecture}

ResNet-50, a variant of ResNet with 50 layers, uses a bottleneck design. The architecture consists of an initial convolutional layer followed by four stages of residual blocks. Each stage contains a different number of bottleneck blocks, with the following configuration:

\[
\begin{array}{c|c|c}
\text{Stage} & \text{Block Type} & \text{Number of Blocks} \\
\hline
1 & 1 \times 1, 64 & 1 \\
2 & \begin{bmatrix} 1 \times 1, 64 \\ 3 \times 3, 64 \\ 1 \times 1, 256 \end{bmatrix} & 3 \\
3 & \begin{bmatrix} 1 \times 1, 128 \\ 3 \times 3, 128 \\ 1 \times 1, 512 \end{bmatrix} & 4 \\
4 & \begin{bmatrix} 1 \times 1, 256 \\ 3 \times 3, 256 \\ 1 \times 1, 1024 \end{bmatrix} & 6 \\
5 & \begin{bmatrix} 1 \times 1, 512 \\ 3 \times 3, 512 \\ 1 \times 1, 2048 \end{bmatrix} & 3 \\
\end{array}
\]

\begin{figure}[h]
 \centering
  \includegraphics[width=.8\linewidth]{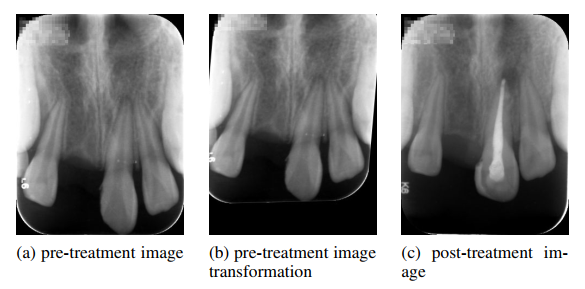}
  \caption{Convolution attention module.}\label{fig.data}
  \vspace{-4mm}
\end{figure}

\subsection{Integration of SimAM}

In this paper, SimAM (Simple Attention Module) is incorporated into each residual block of ResNet-50. SimAM is designed to enhance feature learning by selectively focusing on important regions of the input image. The residual block with SimAM can be expressed as:
\[
\mathbf{y} = \mathcal{F}(\mathbf{x}, \{W_i\}) + \mathbf{x} + \text{SimAM}(\mathbf{x})
\]
where $\text{SimAM}(\mathbf{x})$ represents the attention mechanism applied to the input $\mathbf{x}$.

\begin{table*}[h]
\caption{Result of F1 scores in baseline method.}
\begin{adjustbox}{width=.7\textwidth,center}
\begin{tabular}{|c|c|c|c|c|c|c|}
\hline
Method & Subsampled & Histogram &  PCA & HOG & HOG+PCA  \\ \hline
VGG & 0.434 & 0.535 & 0.455 & 0.556 & 0.637 \\ \hline
EfficientNet & 0.364 & 0.561 & 0.475 & 0.423 & 0.553 \\ \hline
Densnet & 0.471 & 0.543 & 0.542 & 0.625 & \textbf{0.650} \\ \hline
Alexnet & 0.464 & 0.627 & 0.454 & 0.560 & 0.560 \\ \hline
Resnet & 0.347 & 0.580 & 0.440 & 0.327 & 0.577 \\ \hline
Our model & \textbf{0.676} & \textbf{0.647} & \textbf{0.699} & \textbf{0.638} & 0.558 \\ \hline
\end{tabular}
\end{adjustbox}
\end{table*}

\section{Experiments}

\subsection{Dataset}

We organized and interpreted a dataset of 296 pairs of periapical dental radiographs. Each radiograph depicts one or more teeth receiving treatment, with each pair including two radiographs obtained before and after treatment (Figure 2 (a), (c)). Using clinical observations, experienced dentists classified the teeth as 'getting better', 'no change', or 'growing worse'. The radiographs were captured using the following camera settings: 140 kV tube voltage, 80 mA tube current, and a 0.40 msecond exposure duration. During tests, radiologists may make minor changes to these settings. The bisecting angle approach involves varying the shooting angle for each tooth.

To improve the dataset, we used data augmentation. Exposure time and radiation angle can both have an impact on radiography picture quality. To replicate various examination situations, we flipped the photos horizontally and vertically, made incremental rotations, and altered the image brightness within a certain range.

\begin{figure}[h]
 \centering
  \includegraphics[width=.8\linewidth]{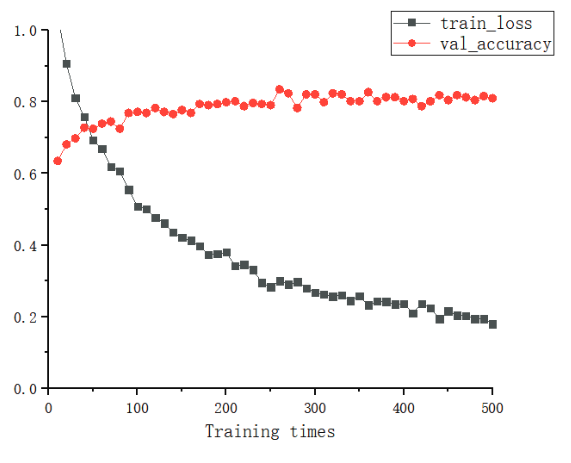}
  \caption{Convolution attention module.}\label{fig.train}
  \vspace{-4mm}
\end{figure}

\subsection{RESULTS}

In this experiment, we conducted training sessions 500 times, and the fluctuations in loss value and accuracy are depicted in Fig 2. It is evident that as the number of training iterations increases, the loss diminishes progressively, while accuracy steadily improves and eventually stabilizes.

The table 1 compares the F1 scores of various models (VGG, EfficientNet, Densnet, Alexnet, Resnet, and "Our model") using different feature extraction techniques (Histogram, Subsampled, HOG, PCA, HOG+PCA). The F1 score is a measure of a test's accuracy and is defined as the harmonic mean of precision and recall. Our model demonstrates robust and consistent performance across multiple feature extraction techniques, indicating its versatility and effectiveness. For example, Our model outperforms all other models with a significant margin, achieving an F1 score of 0.676. The high scores in specific feature extraction techniques suggest potential areas where each model excels and where improvements can be targeted. The choice of feature extraction technique plays a significant role in model performance, with certain combinations (e.g., HOG+PCA) providing notable improvements for specific models. This analysis highlights the strengths and weaknesses of each model and feature extraction technique, providing valuable insights for further optimization and research.

\section{Conclusion}
the integration of deep learning in dental diagnostics presents a significant advancement in enhancing the accuracy and efficiency of disease detection. Our proposed enhanced ResNet50 architecture, incorporating the SimAM attention module, addresses the challenge of limited contrast in dental images and leverages the strengths of deep learning while managing computational demands. The SimAM module effectively captures spatial dependencies and refines feature extraction, leading to improved classification accuracy and robustness in dental image analysis.
The experimental results demonstrate that our model outperforms traditional architectures like VGG, EfficientNet, DenseNet, and AlexNet across various feature extraction techniques. This robust performance highlights the model's versatility and effectiveness in dental applications. Additionally, the comparison of different feature extraction techniques underscores the importance of selecting appropriate methods to enhance model performance.
This advancement not only has the potential to improve patient care and outcomes but also supports the ongoing adoption of AI technologies in dental practices, paving the way for more precise and reliable diagnostic tools in the future.

\bibliographystyle{IEEEtran}

{\small
\bibliography{article}}

\end{document}